\documentclass[conference]{IEEEtran}
\IEEEoverridecommandlockouts
\usepackage{cite}
\usepackage{amsmath,amssymb,amsfonts}
\usepackage{algorithmic}
\usepackage{graphicx}
\usepackage{textcomp}
\usepackage{xcolor}
\usepackage{booktabs}
\usepackage{multirow}
\usepackage{subfigure}
\def\BibTeX{{\rm B\kern-.05em{\sc i\kern-.025em b}\kern-.08em
    T\kern-.1667em\lower.7ex\hbox{E}\kern-.125emX}}
\begin{document}

\title{A New Fine-grained Alignment Method for Image-text Matching\\
}

\author{\IEEEauthorblockN{Yang Zhang}
\IEEEauthorblockA{\textit{School of Electronic Information and Electrical Engineering} \\
\textit{Shanghai Jiao Tong University}\\
Shanghai, China \\
zy1176000t@sjtu.edu.cn}

}

\maketitle

\begin{abstract}
    Image-text retrieval is a widely studied topic in the field of computer vision due to the exponential growth of multimedia data,
    whose core concept is to measure the similarity between images and text. 
    However, most existing retrieval methods heavily rely on cross-attention mechanisms for cross-modal fine-grained alignment,
    which takes into account excessive irrelevant regions and treats prominent and non-significant words equally, thereby limiting retrieval accuracy.
    This paper aims to investigate an alignment approach that reduces the involvement of non-significant fragments in images and text while enhancing the alignment of prominent segments. 
    For this purpose, we introduce the Cross-Modal Prominent Fragments Enhancement Aligning Network(CPFEAN),
    which achieves improved retrieval accuracy by diminishing the participation of irrelevant regions during alignment and relatively increasing the alignment similarity of prominent words. 
    Additionally, we incorporate prior textual information into image regions to reduce misalignment occurrences.
    In practice, we first design a novel intra-modal fragments relationship reasoning method, and subsequently employ our proposed alignment mechanism to compute the similarity between images and text. 
    Extensive quantitative comparative experiments on MS-COCO and Flickr30K datasets demonstrate that our approach outperforms state-of-the-art methods by about 5\% to 10\% in the rSum metric.
\end{abstract}


\section{Introduction}
Recently, the research of cross-modal learning receives wide attention, such as image captioning\cite{b27}, visual question answering\cite{b26}, multimodal image synthesis and editing\cite{b25} and so on.
Image-text retrieval is a fundamental task in cross-modal learning, which retrieves most relevant texts for requested image, and vice versa\cite{b2,b3,b10,b17}.
The primary challenges in image-text retrieval lie in learning image and text feature representations and constructing similarity measurement model. The former involves transforming information from both modalities into vector representations conducive to computation and processing, while the latter enables more precise retrieval of content that meets requirements.

Research in image-text retrieval based on deep learning is primarily divided into two main branches: independent modality learning\cite{b1,b2,b3,b4} and cross-modal fine-grained interaction learning\cite{b7,b8,b9,b10,b11,b23,b24}. 
Independent modality learning embeds images and text into a common feature space.
Initial studies \cite{b1, b2} treated images and text as a whole, leading to embeddings with a substantial amount of redundant information.
As bottom-up attention \cite{b5} has gained recognition, recent research \cite{b3, b4} has started to appreciate the importance of fine-grained fragments.
These studies adopt pooling strategies and self-attention \cite{b6} to fuse features from various regions of the image, 
thereby reducing interference from background information and ultimately enhancing retrieval accuracy. 
However, these studies still do not pay attention to fine-grained interplay between fragments of different modalities, resulting limited retrieval precision.

\begin{figure}[htbp]
    \centerline{\includegraphics[width = 0.45\textwidth]{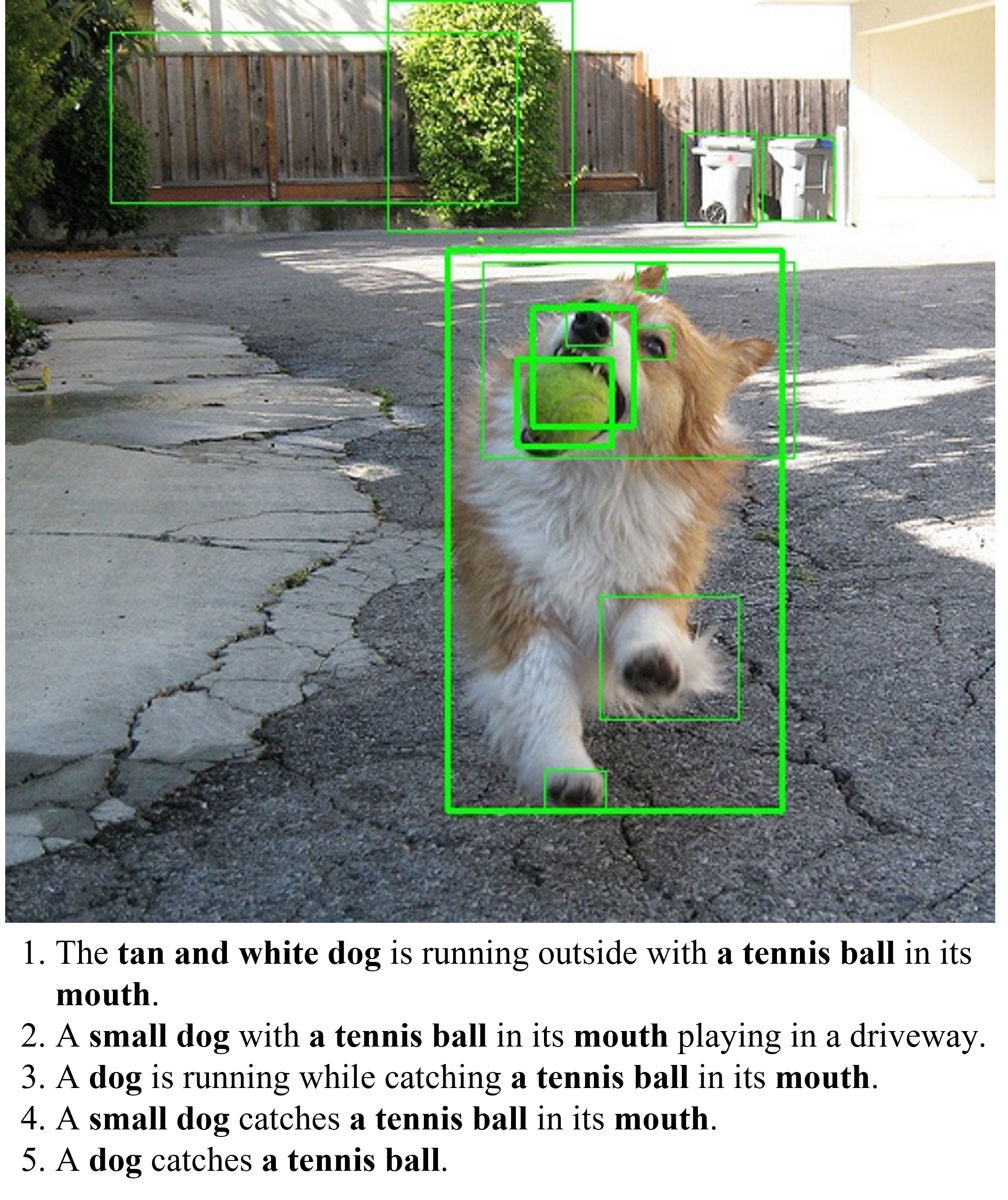}}
    \caption{Prominent fragments in image and text.In the upper image, the green boxes represent the proposed regions extracted by Faster RCNN, with prominent regions outlined in bold. In the lower section, there are five textual descriptions for the image, with recurring prominent words highlighted in bold.}
    \label{fig1}
\end{figure}

Cross-modal fine-grained interaction learning emphasizes the fine alignment between regions and words to discover more accurate semantic associations. 
For instance, SCAN\cite{b7} applies stacked cross-attention to text and image feature representations to create involved vectors, while SGRAF\cite{b8} further incorporates graph convolutional neural networks(GCNs) for fine-grained alignment to form global vectors.
These studies compute the similarity between a fragment of one modality and all fragments of another modality, assigning different weights to all fragments of the other modality. 
This approach prioritizes semantically similar fragments while disregarding less semantically related fragments.

However, on the one hand, we believe that when people describe images, they do not go into great detail but rather focus on the prominent regions. 
Even if non-significant areas of an image change, the text description of that image remains the same. 
As illustrated in Figure \ref{fig1}, many regions are not described in the corresponding text, and these regions become interfering regions in cross-modal alignment. 
SCAN and its variants\cite{b7,b10,b11} consider too many interfering regions during alignment, which limits retrieval accuracy.
On the other hand, people can provide various descriptions for the same image, but prominent words are repeatedly mentioned in these expressions(bold words in Figure \ref{fig1}), while other words can vary. 
From the perspective of words in the text, words highly relevant to regions should receive more attention compared to other words.
Last but not least, during the initial stages of training, the alignment between text and images is relatively chaotic,
which can lead to the misalignment of certain regions during cross-modal alignment, and errors accumulate during training.

To reduce the impact of interference regions during cross-modal alignment, first, we fuse regions with the matching text information for cross-modal semantic enhancement and use a gating mechanism to control information fusion to distinguish between prominent regions and irrelevant regions. 
Second, during alignment, we consider only the most prominent region while ignoring non-significant areas, and the effectiveness of this alignment is also probabilistically explained in CHAN\cite{b19}.
To highlight prominent words in the text, we select the semantically most relevant words from the text when performing cross-modal information fusion with regions, thereby ensuring that prominent words have a higher similarity to the region compared to other words during alignment.
To mitigate the impact of misalignment in the regions, we extract text labels for each region as prior information, enhancing semantic representation consistency.
Alignment similarity scores are increased for regions that are consistent with the prior information semantics, 
while similarity scores for misaligned regions are reduced.

Based on the methods described above, we propose our Cross-modal Prominent Fragments Enhancement Aligning Network(CPFEAN),
which is capable of inferring intra-modal fine-grained semantic associations and performing cross-modal fine-grained alignment.
For fine-grained intra-modal association inference in the image modality, we incorporate prior knowledge by adding textual information extracted from image regions via Faster R-CNN to the image feature representations. 
We employ self-attention to infer semantic relationships and spatial positional relationships among various regions.
For fine-grained intra-modal word association inference in the text modality,  
we employ GCNs to infer attribute relationships among words.
During fine-grained aligning, we align a word from the text with the closest region in the image. 
Additionally, we share information between image regions and cross-modal prominent fragments in the text and use gating mechanism to highlight prominent regions, 
thereby enhancing prominent fragments in both image and text.

Our contributions can be summarized in the following aspects:
\begin{itemize}
\item
We introduce a novel framework for image-text embedding representation that enhances semantic representation consistency between the two modalities to mitigate the effects of misalignment.
\item
We propose a novel cross-modal fine-grained matching strategy that simplifies alignment complexity and improves retrieval accuracy by enhancing prominent fragments alignment.
\item
Our model achieves state-of-the-art results in the field of image-text retrieval. Extensive experiments conducted on the Flickr30K and MS-COCO datasets demonstrate that our model surpasses the previous state-of-the-art models by over 10\% in the rSum metric.
\end{itemize}

\section{Related Works}
\subsection{Independent modal learning}
Independent modality learning aims to map images and text into a common embedding space using two separate networks that do not communicate with each other. 
In early research\cite{b1, b2, b20, b21}, images and text were treated as wholes, with CNNs used to infer image features and RNNs or GRUs used to infer text features. 
Similarity was measured using vector distance. Subsequently, there was a growing focus on extracting region-level features from images. 
\cite{b14} employs scene graphs to construct fine-grained relationships between image regions.
VSE$\infty$\cite{b3} aggregated various image regions into global features using pooling strategies, while VSRN\cite{b4} utilized GCNs to model relationships among region features, 
ultimately summarizing them into global features. With the application of self-attention mechanisms in NLP, 
many studies have gradually adopted BERT\cite{b22} for text feature inference, which inherently emphasizes the mutual relationships among fine-grained text features.

\begin{figure*}[htbp]
    \centerline{\includegraphics[width = \textwidth]{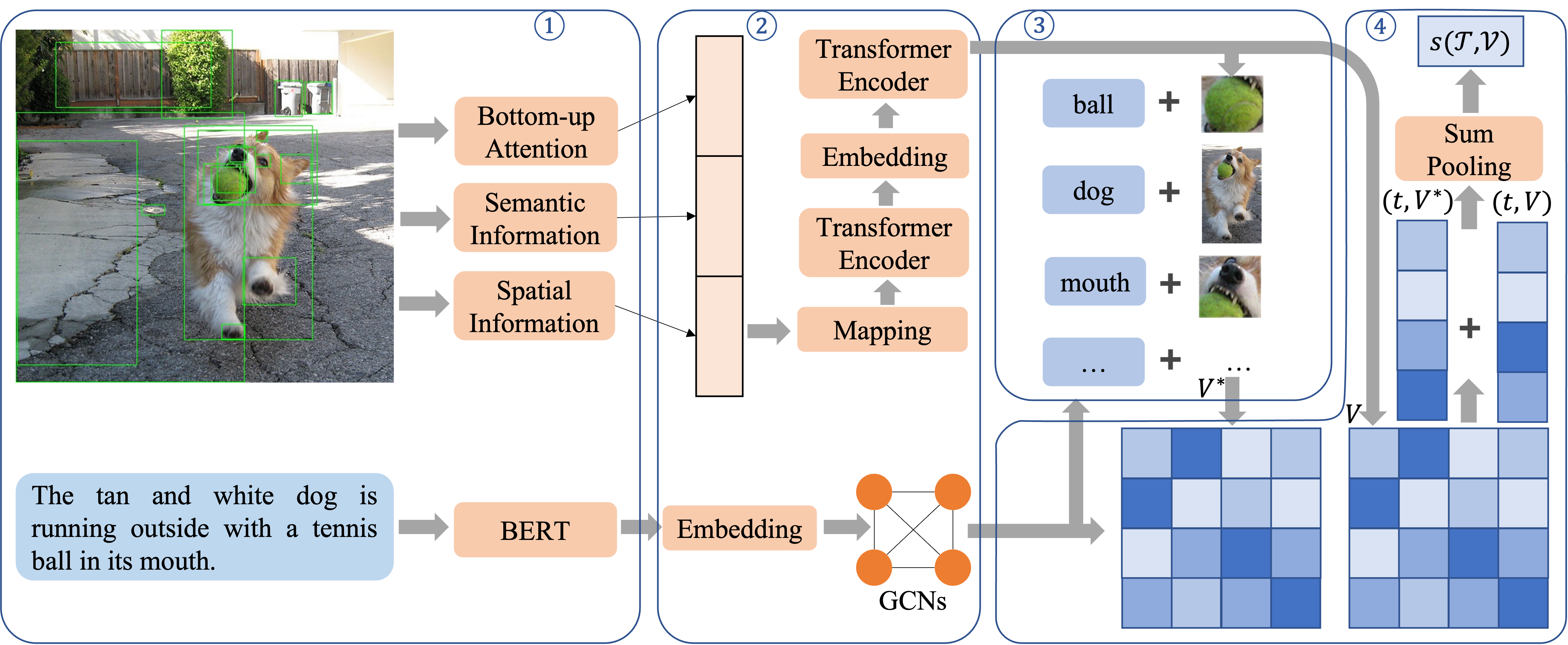}}
    \caption{The workflow of CPFEAN.Our model can be divided into four parts:
    (1)Feature Extraction: This part involves extracting features from both images and text. 
    For the image features, they are comprised of three parts: basic features, spatial features, and semantic features. We use the bottom-up attention features as the basic features, the coordinates and dimensions of each region as spatial features, and the labels extracted by Faster RCNN, processed with a pre-trained language model, as prior semantic features.Text features are extracted using the BERT model.
    (2)Intra-Modal Relationship Reasoning: In this step, we infer the interactions between fragments within each modality. For images, a Transformer encoder is used to reason about the semantic and spatial relationships between regions. For text, a fully connected Graph Convolutional Network is used to infer the semantic relationships between words.
    (3)Cross-Modal Semantic Fusion: This stage involves fusing the regions of the image separately with the cross-modal prominent fragments in the text, resulting in image representations with cross-modal semantic features (denoted as $V^*$).
    (4)Cross-Modal Prominent Fragments Alignment: In this part, each word is aligned separately with the cross-modal prominent fragments in both the $V$ and $V^*$ representations of the image. The similarity scores for these alignments are summed to compute the similarity between each word and the image. Finally, a summation pooling operation yields the text-image similarity.}
    \label{fig2}
\end{figure*}

\subsection{Cross-modal fine-grained interaction learning}
Cross-modal fine-grained interaction learning involves aligning fine-grained fragments from both modalities 
in image and text representations or when measuring similarity. This alignment leverages fine-grained semantic 
relationships to achieve more accurate similarity measurements. SCAN\cite{b7} employs stacked cross-attention to calculate 
similarity between individual fragments and the overall fragments. IMRAM\cite{b11} builds upon this by introducing memory distillation 
units and iteratively extracting cross-modal information. Many studies\cite{b8,b12,b13,b16,b17} incorporate GCNs into fine-grained alignment. 
For example, CGMN\cite{b12} employs graph convolutional networks to infer feature representations for both images and text. During training, 
it adds fine-grained alignment between the two graphs as an additional loss. On the other hand, 
CSMEI\cite{b17} utilizes scene graphs to construct visual semantic graphs, which are jointly integrated with visual spatial graphs to create a comprehensive visual graph. 
Finally, fine-grained alignment is performed with the text graph. 
Additionally, NAAF\cite{b9} enhances the discriminative and robust qualities of negative effects by mining mismatched fragments.

\section{Cross-modal Prominent Fragments Enhancement Aligning Network}
This section provides a detailed overview of our Cross-modal Prominent Fragments Enhancement Aligning Network, as depicted in Figure \ref{fig2}.
Firstly, Section \ref{31} introduces the representations of text and image. In Section \ref{32}, 
we delve into the inference of fragment relationships within each modality. 
Following that, Section \ref{33} explains the cross-modal prominent fragments enhancement alignment, 
and finally, Section \ref{34} discusses the loss functions.

\subsection{Image and Text Representation}\label{31}
\textbf{Image representation.} We utilize a bottom-up attention model\cite{b5} to extract features from $m$ salient regions of an image. 
Specially, given an image $\mathcal{V}$, the region features can be denoted as $I = \{r_1, r_2, r_3, ..., r_m\}$, $r_i \in \mathbb{R}^{2048}$.
These initial region features are further enhanced with spatial and semantic information in subsequent steps.

\textbf{Text representation.} We employ the well-performing pre-trained BERT\cite{b22} model from the field of NLP to extract features for a text.
Specially, consider a text $\mathcal{T}$ consisting of $n$ words, the word features can be denoted as $S = \{w_1, w_2, w_3, ..., w_n\}$, $w_i \in \mathbb{R}^{768}$.
Similar to the region features, these word features also serves as initial information awaiting further processing.

\subsection{Intra-modal Fragment Relationships Reasoning}\label{32}
For image regions, we consider them to have both spatial positional relationships and semantic relationships. 
Unlike many studies\cite{b17,b28} that use scene graphs to represent semantic relationships, 
we use the labels extracted by object detection networks as textual information to construct semantic relationships. 
Finally, we employ a Transformer to infer the ultimate representation of the regions.
As for the words in the text, we create a fully connected text graph and employ a graph convolutional neural network 
to infer the semantic relationships between words.
\subsubsection{Visual Semantic and Spatial Relationships Reasoning}
To mitigate the accumulation of training errors resulting from inaccurate textual descriptions of images in the dataset, 
we propose introducing text prior information into the image representation.
We employ the textual information extracted from region proposals by Faster R-CNN as semantic features. 
To maintain semantic consistency, we also use pre-trained BERT to extract features for individual words in the region's text, 
and use max-pooling to generate an overall semantic feature $rt_i = max\{rw_1, rw_2, ...\},rw_i \in \mathbb{R}^{768}$ for $i$th region. 
Then, we use the region's position as spatial feature $rs_i$.
\begin{equation}
    rs_i = \{\frac{x_1}{w}, \frac{y_1}{h}, \frac{x_2}{w}, \frac{y_2}{h}, \frac{(x_2 - x_1)}{w}, \frac{(y_2 - y_1)}{h} \}
\end{equation}
\noindent where $(x_1, y_1)$,$(x_2, y_2)$ are the coordinates of the top-left and bottom-right corners of the region, and $w$, $h$ are the width and height of the image.

We concatenate the initial region features with the semantic and spatial features and then 
employ an FC-ReLU-FC module to map the visual features to a 2048-dimensional space, 
now we get $\hat{I} = \{\hat{r_1}, \hat{r_2}, ..., \hat{r_m}\},\hat{r_i} \in \mathbb{R}^{2048}$. The above process can be described as follows:
\begin{equation}
    \hat{r_i} = W_v(ReLU(W_r\{r_i, rs_i, rt_i\}))
\end{equation}
\noindent where $W_r$ and $W_v$ are weights and bias, and share weights for all of $m$ regions.

Lastly, we utilize a Transformer layers-linear projection layers-Transformer layers to infer relationships between various regions in the image. 
The linear projection layer maps the image to a $D$-dimensional embedding space, facilitating fine-grained alignment with the text.
In the final inference step of the Transformer, we obtain the ultimate representation of the image, 
denoted as $V = \{v_1, v_2, ..., v_m\},v_i \in \mathbb{R}^D$.
\subsubsection{Textual Semantic Relationships Reasoning}
Similar to image processing, we embed the initial features of words $S$ into the same D-dimensional feature space using a fully connected layer, resulting in $\hat{S} = \{\hat{w_1}, \hat{w_2}, ..., \hat{w_n} \}, \hat{w_i} \in \mathbb{R}^D$.
Many studies\cite{b12,b13} construct sparse graphs using the syntactic dependency matrices generated by Stanford CoreNLP\cite{b29} when building text graphs. 
In contrast, we directly build dense graphs. We believe that fully connected graphs can provide a more comprehensive analysis of the dependencies between words, 
while maintaining relatively lower model complexity.
Similar to reference\cite{b4,b17,b30}, we construct a pairwise affinity matrix $R$ between embedding vectors:
\begin{equation}
    R(\hat{w_i},\hat{w_j}) = (W_\phi \hat{w_i})^T(W_\varphi \hat{w_j})
\end{equation}
\noindent where $W_\phi$ and $W_\varphi$ are two embedding matrix of dimension $D \times D$ with learnable weight parameters.

Subsequently, we use the embedding vectors as nodes and the affinity matrix as edges to construct a dense graph, and employ GCNs to infer the semantic relationships between words. 
We connect the feature vectors generated by GCNs and the original embedding vectors through residual connections,
denoted as $T = \{t_1, t_2, ..., t_n\}$, as follows:
\begin{equation}
    T = W_r(R\hat{S}W_g) + \hat{S}
\end{equation}
\noindent where $W_r$ represents the weight matrix for the GCNs layer, $W_g$ is the weight matrix for the residual structure, and $R$ is the affinity matrix.

\subsection{Cross-modal Prominent Fragments Enhancement Alignment}\label{33}
After performing intra-modality relationship inference, we obtain the final representations of text and image features. 
Now, we construct a similarity measurement model between text and image. 
We perform cross-modal fine-grained alignment of text and image, enhancing pairs of fragments with stronger semantic correlations and 
weakening pairs with weaker semantic relationships to more accurately measure the similarity between text and image.
\subsubsection{Semantic fusion using cross-modal prominent fragments}
For a fragment in one modality(region or word), we refer to the fragment in the other modality(word or region) 
that is semantically closest to it as its cross-modal prominent fragment.
We utilize cosine distance to measure semantic distance.
For instance, given a query region $v_i$, we obtain its cross-modal prominent fragment $t_k$ as follows:
\begin{equation}
    \begin{aligned}
        k &= \mathop{\mathrm{argmax}}\limits_{j = 1\dots n}{s(v_i,t_j)} \\
        s&(v_i,t_j) = \frac{{v_i}^Tt_j}{\|v_i\|\|t_j\|}
    \end{aligned}
\end{equation}

We believe that if the textual description of the image is accurate, $t_k$ is the most appropriate cross-modal semantic representation for $v_i$. 
Hence, we use cross-modal prominent fragments to enhance semantics. 
In order to emphasize prominent regions and reduce the impact of interfering regions in cross-modal fine-grained alignment, 
we use a gating mechanism\cite{b11} to control semantic fusion:
\begin{equation}
    \begin{aligned}
        g_i &= \sigma(W_g\{v_i,t_k\}) \\
        {v_i}^* &= g_i * v_i + (1 - g_i)*\tanh(W_h\{v_i,t_k\})
    \end{aligned}
\end{equation}
where $W_g$ and $W_h$ are weights and bias, and share weights for all of words.

Thus, we obtain the representation $V^* = \{{v_1}^*, {v_2}^*, ..., {v_n}^*\}$ with cross-modal semantic enhancement. 
\subsubsection{Cross-modal prominent fragments alignment}
When performing fine-grained semantic alignment, we treat the similarity between a word and its cross-modal prominent fragment 
as the similarity between the word and the entire image, while ignoring all other cross-modal fragments. 
We take into account the similarity between a word and the representation $V$ with basic semantic information and 
the representation $V^*$ with advanced cross-modal semantic information, as follows:
\begin{equation}
    \begin{aligned}
        s(t_i, \mathcal{V}) &= s(t_i, V) + s(t_i, V^*) \\
        &= \mathop{\mathrm{max}}\limits_{j = 1\dots m}{s(t_i,v_j)} + \mathop{\mathrm{max}}\limits_{j = 1\dots m}{s(t_i,{v_j}^*)}
    \end{aligned}
\end{equation}

Lastly, we sum the similarities between all words and the image to represent the overall similarity between the text and the image:
\begin{equation}
    s(\mathcal{T}, \mathcal{V}) = \sum_{i=1}^{n}s(t_i, \mathcal{V})
\end{equation}

\begin{table*}[htbp]
    \caption{Comparison of experimental results on MS-COCO 5-fold 1K test set and 5K test set}
    \setlength\tabcolsep{12pt}
    \begin{center}
    \begin{tabular}{*{8}{c}}
        \toprule
        \multirow{2}*[-0.5ex]{\textbf{Method}} & \multicolumn{3}{c}{\textbf{Text Retrieval}} & \multicolumn{3}{c}{\textbf{Image Retrieval}} & \multirow{2}*[-0.5ex]{\textbf{rSum}} \\
        \cmidrule(lr){2-4}\cmidrule(lr){5-7} 
        & R@1 & R@5 & R@10 & R@1 & R@5 & R@10 \\
        \midrule
        \multicolumn{8}{c}{\textbf{COCO 5-fold 1K Test}} \\
        \midrule
        VSE++$_{2017}$\cite{b2} & 64.6 & 90.0 & 95.7 & 52.0 & 84.3 & 92.0 & 478.6 \\
        SCAN$^*_{2018}$\cite{b7} & 72.7 & 94.8 & 98.4 & 58.8 & 88.4 & 94.8 & 507.9 \\
        VSRN$^*_{2019}$\cite{b4} & 76.2 & 94.8 & 98.2 & 62.8 & 89.7 & 95.1 & 516.8 \\
        IMRAM$^*_{2020}$\cite{b11} & 76.7 & 95.6 & 98.5 & 61.7 & 89.1 & 95.0 & 516.6 \\
        CAAN$^*_{2020}$\cite{b33} & 75.5 & 95.4 & 98.5 & 61.3 & 89.7 & 95.2 & 515.6 \\
        GSMN$^*_{2020}$\cite{b13} & 78.4 & 96.4 & 98.6 & 63.3 & 90.1 & 95.7 & 522.5 \\
        SGRAF$^*_{2021}$\cite{b8} & 79.3 & 96.7 & 98.3 &  64.5 &  90.0 & 95.8 & 524.6 \\
        VSE$\infty$$_{2021}$\cite{b3} & 79.7 & 96.4 & 98.9 & 64.8 & 91.4 & 96.3 & 527.5 \\
        TERAN$^*_{2021}$\cite{b15} & 80.2 & 96.6 & 99.0 & 67.0 & 92.2 & 96.9 & 531.9 \\
        CGMN$_{2022}$\cite{b12} & 76.8 & 95.4 & 98.3 & 63.8 & 90.7 & 95.7 & 520.7 \\
        NAAF$_{2022}$\cite{b9} & 78.1 & 96.1 & 98.6 & 63.5 & 89.6 & 95.3 & 521.2 \\
        VSRN++$^*_{2022}$\cite{b30} & 77.9 & 96.0 & 98.5 & 64.1 & 91.0 & 96.1 & 523.6 \\
        CHAN$_{2023}$\cite{b19} & 81.4 & 96.9 & 98.9 & 66.5 & 92.1 & 96.7 & 532.6 \\
        CMSEI$^*_{2023}$\cite{b17} & 81.4 & 96.6 & 98.8 & 65.8 & 91.8 & 96.8 & 531.1 \\
        HREM$^*_{2023}$\cite{b16} & \textbf{82.9} & 96.9 & 99.0 & 67.1 & 92.0 & 96.6 & 534.6 \\
        \midrule
        CPFEAN(ours) & 81.8 & \textbf{97.3} & \textbf{99.2} & \textbf{69.9} & \textbf{93.7} & \textbf{97.4} & \textbf{539.3}\\
        \midrule
        \multicolumn{8}{c}{\textbf{COCO 5K Test}} \\
        \midrule
        VSE++$_{2017}$\cite{b2} & 41.3& 71.1& 81.2& 30.3& 59.4& 72.4& 355.7 \\
        SCAN$^*_{2018}$\cite{b7} & 50.4& 82.2& 90.0& 38.6& 69.3& 80.4& 410.9 \\
        VSRN$^*_{2019}$\cite{b4} & 53.0& 81.1& 89.4& 40.5& 70.6& 81.1& 415.7 \\
        IMRAM$^*_{2020}$\cite{b11} & 53.7& 83.2& 91.0& 39.7& 69.1& 79.8& 416.5 \\
        CAAN$^*_{2020}$\cite{b33} & 52.5 & 83.3 & 90.9 & 41.2 & 70.3 & 82.9 & 421.1 \\
        SGRAF$^*_{2021}$\cite{b8} & 55.8& 83.0& 91.0& 42.0& 72.4& 82.1& 426.3 \\
        VSE$\infty$$_{2021}$\cite{b3} & 56.6& 83.6& 91.4& 39.3& 69.9& 81.1& 421.9 \\
        TERAN$^*_{2021}$\cite{b15} & 59.3& 85.8& 92.4& 45.1& \textbf{76.4}& 84.4& 443.4 \\
        CGMN$_{2022}$\cite{b12} & 53.4& 81.3& 89.6& 41.2& 71.9& 82.4& 419.8 \\
        NAAF$_{2022}$\cite{b9} & 58.9& 85.2& 92.0& 42.5& 70.9& 81.4& 430.9 \\
        VSRN++$^*_{2022}$\cite{b30} & 54.7& 82.9& 90.9& 42.0 &72.2 &82.7& 425.4 \\
        CHAN$_{2023}$\cite{b19} & 59.8& 87.2& 93.3& 44.9& 74.5& 84.2& 443.9 \\
        CMSEI$^*_{2023}$\cite{b17} & 61.5 & 86.3 & 92.7 & 44.0 & 73.4 & 83.4 & 441.2 \\
        HREM$^*_{2023}$\cite{b16} & \textbf{64.0} & \textbf{88.5} & 93.7 & 45.4 & 75.1 & 84.3 & 450.9 \\
        \midrule
        CPFEAN(ours) & 61.6 & 87.0 & \textbf{93.7} & \textbf{47.8} & 76.2 & \textbf{85.8} & \textbf{452.0}\\
        \bottomrule
    \end{tabular}
    \label{tabcoco}
    \end{center}
\end{table*}

\subsection{Objective Function}\label{34}
A bi-directional ranking triplet loss\cite{b2} with hard negative samples is employed as the objective function to optimize 
the parameters of the entire network. This aims to maximize the distance between positive sample 
pairs and hard negative sample pairs within a batch.The objective function is defined as:
\begin{equation}
    \begin{aligned}
        \mathcal{L} = \sum_{(\mathcal{T},\mathcal{V}) \in B}{}[\alpha - s(\mathcal{T},\mathcal{V}) + s(\mathcal{T},\mathcal{V}^-)]_+& \\
        + [\alpha - s(\mathcal{T},\mathcal{V}) + s(\mathcal{T}^-,\mathcal{V})]_+&
    \end{aligned}
\end{equation}
where $\alpha$ is a margin parameter, $[x]_+ = max(x, 0)$, $(\mathcal{T},\mathcal{V})$ is a positive text-image pair 
and $(\mathcal{T},\mathcal{V}^-),(\mathcal{T}^-,\mathcal{V})$ are hard negative pairs in a mini batch $B$.

\section{Experiments}
\subsection{Datasets and Evaluation Metrics}
\subsubsection{Datasets}We evaluate our model using two datasets, MS-COCO\cite{b31} and Flickr30K\cite{b32}. MS-COCO contains 123,287 images, 
each with five corresponding textual descriptions. Following \cite{b2,b15}, we use 113,287 images for training, 
5,000 images for validation, and the remaining 5,000 images for testing. The test results for MS-COCO consist of the averaged 
results from a five-fold cross-validation on 1,000 test samples (COCO 5-fold 1k test) and the complete 5,000 test samples (COCO 5k test). 
Flickr30K comprises 31,783 images, each with five textual descriptions. Following the split in \cite{b2}, we use 1,014 images for validation, 
1,000 images for testing, and 29000 images for training.
\subsubsection{Evaluation Metrics}Text-image retrieval is typically evaluated using Recall@K (K=1, 5, 10) metrics, denoted as R@1, R@5, and R@10. 
Recall@K represents the percentage of ground truth among the top K retrieved items, with higher values indicating higher retrieval accuracy. 
We compute the sum of the three image retrieval metrics and the sum of the three text retrieval metrics, and combine them as the total evaluation metric for text-image retrieval, referred to as rSum:
\begin{equation}
    \begin{aligned}
        rSum = \underbrace{R@1 + R@5 + R@10} + &\underbrace{R@1 + R@5 + R@10} \\
        Image \enspace Retrieval \quad  & \quad \quad Text \enspace Retrieval
    \end{aligned}
\end{equation}

\begin{table*}[htbp]
    \caption{Comparison of experimental results on Flickr30K 1K test set}
    \setlength\tabcolsep{12pt}
    \begin{center}
        \begin{tabular}{*{8}{c}}
            \toprule
            \multirow{2}*[-0.5ex]{\textbf{Method}} & \multicolumn{3}{c}{\textbf{Text Retrieval}} & \multicolumn{3}{c}{\textbf{Image Retrieval}} & \multirow{2}*[-0.5ex]{\textbf{rSum}} \\
            \cmidrule(lr){2-4}\cmidrule(lr){5-7} 
            & R@1 & R@5 & R@10 & R@1 & R@5 & R@10 \\
            \midrule
            \underline{VSE++}$_{2017}$\cite{b2} & 52.9 &80.5 &87.2 &39.6 &70.1 &79.5 &409.8 \\
            \underline{SCAN}$^*_{2018}$\cite{b7} & 67.4 &90.3 &95.8 &48.6 &77.7 &85.2 &465.0 \\
            \underline{VSRN}$^*_{2019}$\cite{b4} & 71.3 &90.6 &96.0 &54.7 &81.8 &88.2 &482.6 \\
            \underline{IMRAM}$^*_{2020}$\cite{b11} & 74.1 &93.0 &96.6 &53.9 &79.4 &87.2 &484.2 \\
            \underline{CAAN}$^*_{2020}$\cite{b33} & 70.1 & 91.6 & 97.2 & 52.8 & 79.0 & 87.9 & 478.6 \\
            \underline{GSMN}$^*_{2020}$\cite{b13} & 76.4 & 96.3 & 97.3 & 57.4 & 82.3 & 89.0 & 496.8 \\
            \underline{SGRAF}$^*_{2021}$\cite{b8} & 78.4 &94.6 &97.5 &58.2 &83.0 &89.1 &500.8 \\
            VSE$\infty$$_{2021}$\cite{b3} & 76.5 &94.2 &97.7 &56.4 &83.4 &89.9 &498.1 \\
            TERAN$^*_{2021}$\cite{b15} & 79.2 &94.4 &96.8 &63.1 &87.3 &92.6 &513.4 \\
            \underline{CGMN}$_{2022}$\cite{b12} & 77.9 &93.8 &96.8 &59.9 &85.1 &90.6 &504.1 \\
            \underline{NAAF}$_{2022}$\cite{b9} & 79.6 &96.3 &98.3 &59.3 &83.9 &90.2 &507.6 \\
            VSRN++$^*_{2022}$\cite{b30} & 79.2 &94.6 &97.5 &60.6 &85.6 &91.4 &508.9 \\
            CHAN$_{2023}$\cite{b19} & 80.6 &96.1 &97.8 &63.9 &87.5 &92.6 &518.5 \\
            CMSEI$^*_{2023}$\cite{b17} & 82.3 & 96.4 & 98.6 & 64.1 & 87.3 & 92.6 & 521.3 \\
            HREM$^*_{2023}$\cite{b16} & \textbf{84.0} & 96.1 & 98.6 & 64.4 & 88.0 & 93.1 & 524.2 \\
            \midrule
            CPFEAN(ours) & 83.2 & \textbf{97.1} & \textbf{98.9} & \textbf{69.4} & \textbf{91.0} & \textbf{95.1} & \textbf{534.7}\\
            \bottomrule
        \end{tabular}
        \label{tabf30k}
    \end{center}
\end{table*}

\subsection{Implementation Details}
For text encoding, we fine-tune the pre-trained BERT model on the titles of both datasets. For image encoding, the number of extracted regions differs from most other research. 
SCAN and its variants\cite{b7,b10,b11} consider too many redundant regions during alignment, resulting in decreased performance when the number of regions exceeds 36. However, our method doesn't exhibit the same behavior. 
Therefore, we retain all proposed bounding boxes with an IoU over 0.2, while other settings remain the same as bottom-up attention\cite{b5}.
The embedding space dimensions for both images and text are set to $D = 1024$.

Our experiments were conducted on an NVIDIA GTX 1080Ti, and the model was implemented in PyTorch. During training, we used the Adam optimizer with an initial learning rate of 1e-5, which was decayed by a factor of 0.1 every 15 epochs. 
The batch size was set to 16, and the margin parameter $\alpha$ was set to 0.2.

\subsection{Comparison Results with State-of-th-art Methods}
We have compared our proposed CPFEAN method with the state-of-the-art approaches in recent years. All the methods considered for comparison, except for VSE++\cite{b2}, rely on bottom-up attention mechanisms for image feature extraction 
and utilize either Bi-GRU or BERT for text feature extraction. In cases where a method presents results for both Bi-GRU and BERT based text feature extraction, 
we have exclusively shown the results based on BERT. The methods solely based on Bi-GRU is indicated with an underscore. 
The best results for Recall@K and rSum among all methods are highlighted in bold. Methods marked with an asterisk (*) indicate that they employ an ensemble of two models, 
while our method does not utilize such model ensemble.
\subsubsection{Quantitative comparison on MS-COCO}
Table 1 provides a comparison of our results on the MS-COCO dataset with state-of-the-art methods in recent years. 
In the COCO 5-fold 1K test set, our model achieves the highest performance in 5 out of 6 R@K metrics, with the remaining one ranking second. 
Our model outperforms recent methods like CHAN\cite{b19}, CMSEI\cite{b17}, and HREM\cite{b16} by 6.7\%, 8.2\%, and 4.7\%, respectively, in terms of rSum. 
While our model slightly lags behind HREM\cite{b16} in text retrieval, it shows significant improvement in image retrieval.

On the COCO 5k test set, our model also demonstrates a state-of-the-art performance. 
Out of the 6 R@K metrics, we still have 3 achieving the highest level, while the remaining two are ranked second, third, and second, respectively. 
Regarding the rSum metric, we outperform HREM\cite{b16}, CHAN\cite{b19}, and TERAN\cite{b15} by 1.1\%, 8.1\%, and 8.6\%, respectively. 
Our model continues to make significant strides in image retrieval, thanks to our cross-modal prominent fragments enhancement alignment method.

\subsubsection{Quantitative comparison on Flickr30K}
Table \ref{tabf30k} presents the quantitative comparison results of all methods on the Flickr30K 1K test set. 
Compared to the MS-COCO dataset, our method exhibits a more substantial improvement on the Flickr30K dataset.
Our approach achieved state-of-the-art performance in all metrics, except for the Recall@1 in text retrieval, where it ranked second.
Our approach particularly excelled in the three metrics of image retrieval, surpassing the second-ranked method by 5\%, 3\%, and 2\%, respectively. 
In text retrieval, our results were comparable to HREM\cite{b16}, with minimal differences in the three individual metrics, but the total of them continued to demonstrate state-of-the-art performance. 
Overall, our rSum outperformed the second-ranked method by 10.5\%.

\subsection{Ablation Studies}
We conducted ablation experiments on the Flickr30K dataset to analyze the effectiveness of each component of our model. 
For a more comprehensive comparison, we simply replaced cross-attention mechanism of SCAN\cite{b7} and IMRAM\cite{b11} with our proposed prominent fragment enhancement alignment method, 
and performed comparative experiments with text feature extraction based on BiGRU.
\subsubsection{Effects of cross-modal semantic fusion}
We removed the Cross-Modal Semantic Fusion (CSF) module and only considered the image representation $V$ with basic semantic information for fine-grained alignment during experiments. 
The results, as shown in Table \ref{tabnet}, indicate that the removal of the CSF module led to a significant decrease in both image and text retrieval accuracy. 
Text retrieval R@1 decreased by 1.9\%, image retrieval R@1 decreased by 2.9\%, and the rSum metric decreased by 5.9\%. 
The impact of cross-modal semantic fusion is more pronounced in image retrieval compared to text retrieval, possibly because the number of interfering regions in images is larger than the number of unimportant words in text, 
and therefore has a greater influence on retrieval, as text descriptions are generally more concise and accurate.
\subsubsection{Effects of prior text information}
We conducted experiments to evaluate the impact of removing Prior Textual Information (PTI) from the model while keeping all other components unchanged. 
The results showed a significant decrease in all six metrics, with a 7.2\% drop in the rSum metric, indicating the importance of prior information in image representation. 
Furthermore, we conducted a comparative experiment in which both CSF and PTI were removed simultaneously. 
The results indicated that removing both components led to a 9.6\% drop in rSum. However, compared to removing either CSF or PTI alone, the accuracy only dropped by approximately 3\%. 
This suggests that prior textual information and cross-modal semantic fusion complement each other. 
We speculate that the introduction of prior information helps the fusion process more accurately identify cross-modal prominent fragments, thereby enhancing the overall accuracy.
\subsubsection{Effects of textual graph reasoning}
To demonstrate the effectiveness of word semantic relationship reasoning, 
we removed the Textual Graph Reasoning module (TGR) and directly mapped the vectors from BERT to the embedding space. 
Experimental results showed a significant decrease in retrieval accuracy after removing TGR, with an 8.0\% drop in the rSum metric. 
The drop was especially pronounced in image retrieval accuracy, highlighting the importance of modality-specific relationship reasoning. 
We also compared the use of a Transformer encoder (TTE) to infer word relationships instead of GCNs. 
The experimental results indicated that TTE was less effective in reasoning compared to TGR, 
resulting in decreased accuracy for both image and text retrieval.
\subsubsection{Effects of prominent fragments enhancement alignment}
To demonstrate the effectiveness of our proposed prominent fragment enhancement alignment approach, we compared it with other fine-grained alignment methods. 
We selected the same text and image representation methods as SCAN\cite{b7} and IMRAM\cite{b11}, where text features are extracted using BiGRU, image features are extracted using BUTD, and both of them are embedded into the same-dimensional feature space. 
Table \ref{tabgru} presents the results comparing the three fine-grained alignment methods.

For SCAN, we selected two alignment modes, i-t and t-i, where i-t had higher retrieval accuracy. Compared to SCAN i-t, our method showed a significant improvement in retrieval accuracy, with a 23.8\% increase in the rSum metric. 
This improvement is more pronounced in image retrieval relative to text retrieval, consistent with the conclusion drawn from Table \ref{tabnet}. 
As for IMRAM, we also selected two alignment modes, where Text-IMRAM showed higher retrieval accuracy. Compared to Text-IMRAM, our experimental results were on par, but our model had shorter retrieval times 
because we only considered semantically closest segments during alignment and fusion, and we performed fusion only once. 
We also conducted experiments with two rounds of cross-modal semantic fusion, matching Text-IMRAM in the number of fusions.
The results showed that our method had a significant improvement, with a 5.4\% increase in the rSum metric, demonstrating that our approach achieves higher retrieval accuracy in a shorter retrieval time.

\begin{table}[htbp]
    \caption{Ablation studies on Flickr30K 1K test set}
    \setlength\tabcolsep{3pt}
    \begin{center}
        \begin{tabular}{*{8}{c}}
            \toprule
            \multirow{2}*[-0.5ex]{\textbf{Method}} & \multicolumn{3}{c}{\textbf{Text Retrieval}} & \multicolumn{3}{c}{\textbf{Image Retrieval}} & \multirow{2}*[-0.5ex]{\textbf{rSum}} \\
            \cmidrule(lr){2-4}\cmidrule(lr){5-7} 
            & R@1 & R@5 & R@10 & R@1 & R@5 & R@10 \\
            \midrule
            w/o CSF & 81.3 & \textbf{97.4} & \textbf{99.2} & 66.5 & 89.8 & 94.6 & 528.8 \\
            w/o PTI & 82.1 & 96.3 & 98.8 & 66.7 & 89.5 & 94.2 & 527.5 \\
            w/o CSF \& PTI &  80.3 & 96.9 &98.8 & 66.0 & 89.0 & 94.1 & 525.1 \\
            w/o TGR & 82.3 & 96.0 & 98.1 & 66.9 & 89.2 & 94.2 & 526.7 \\
            w. TTE(TGR) & 82.7 & 97.7 & 98.9 & 67.8 & 90.1 & 94.5 & 531.8 \\
            \midrule
            CPFEAN & \textbf{83.2} & 97.1 & 98.9 & \textbf{69.4} & \textbf{91.0} & \textbf{95.1} & \textbf{534.7}\\
            \bottomrule
        \end{tabular}
        \label{tabnet}
    \end{center}
\end{table}

\begin{table}[htbp]
    \caption{Comparison of fine-grained alignment on Flickr30K 1K test set}
    \setlength\tabcolsep{3pt}
    \begin{center}
        \begin{tabular}{*{8}{c}}
            \toprule
            \multirow{2}*[-0.5ex]{\textbf{Method}} & \multicolumn{3}{c}{\textbf{Text Retrieval}} & \multicolumn{3}{c}{\textbf{Image Retrieval}} & \multirow{2}*[-0.5ex]{\textbf{rSum}} \\
            \cmidrule(lr){2-4}\cmidrule(lr){5-7} 
            & R@1 & R@5 & R@10 & R@1 & R@5 & R@10 \\
            \midrule
            SCAN i-t & 67.9 & 89.0 & 94.4 & 43.9 & 74.2 & 82.8 & 452.2 \\
            SCAN t-i & 61.8 & 87.5 & 93.7 & 45.8 & 74.4 & 83.0 & 446.2 \\
            Image-IMRAM & 67.0 & 90.5 & 95.6 & 51.2 & 78.2 & 85.5 & 468.0 \\
            Text-IMRAM &  68.8 & 91.6 &96.0 & 53.0 & 79.0 & \textbf{87.1} & 475.5 \\
            \midrule
            CSF & 70.7 & 90.7 & 95.8 & 52.4 & \textbf{79.7} & 86.8 & 476.0 \\
            CSF twice & \textbf{73.0} & \textbf{92.5} & \textbf{96.3} & \textbf{52.6} & 79.5 & 86.9 & \textbf{480.9} \\
            \bottomrule
        \end{tabular}
        \label{tabgru}
    \end{center}
\end{table}

\begin{figure*}[htbp]
    \centering
    \subfigure[Alignment before semantic fusion.]{\includegraphics[width=0.33\textwidth]{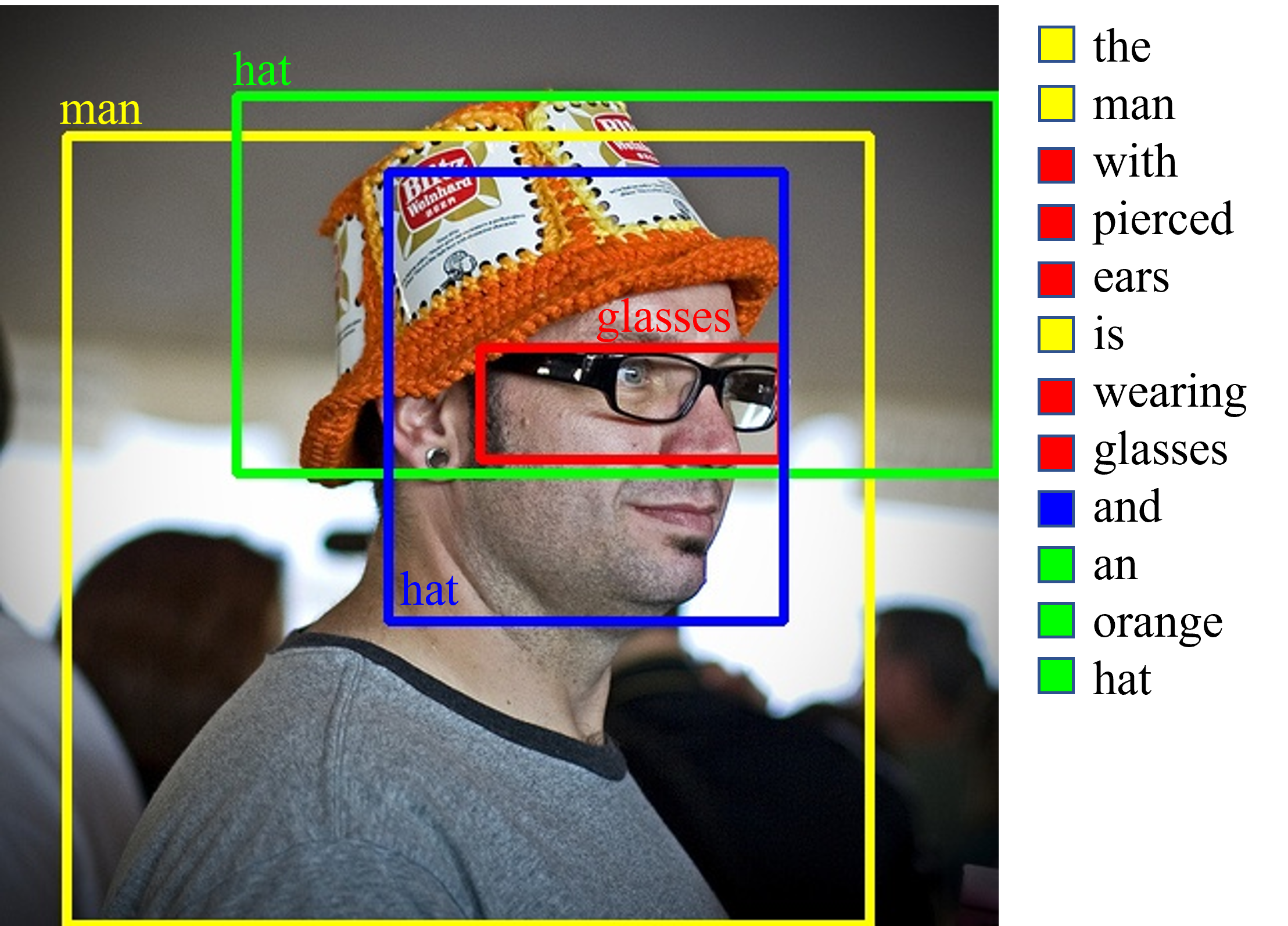}}
    \subfigure[Alignment after semantic fusion.]{\includegraphics[width=0.33\textwidth]{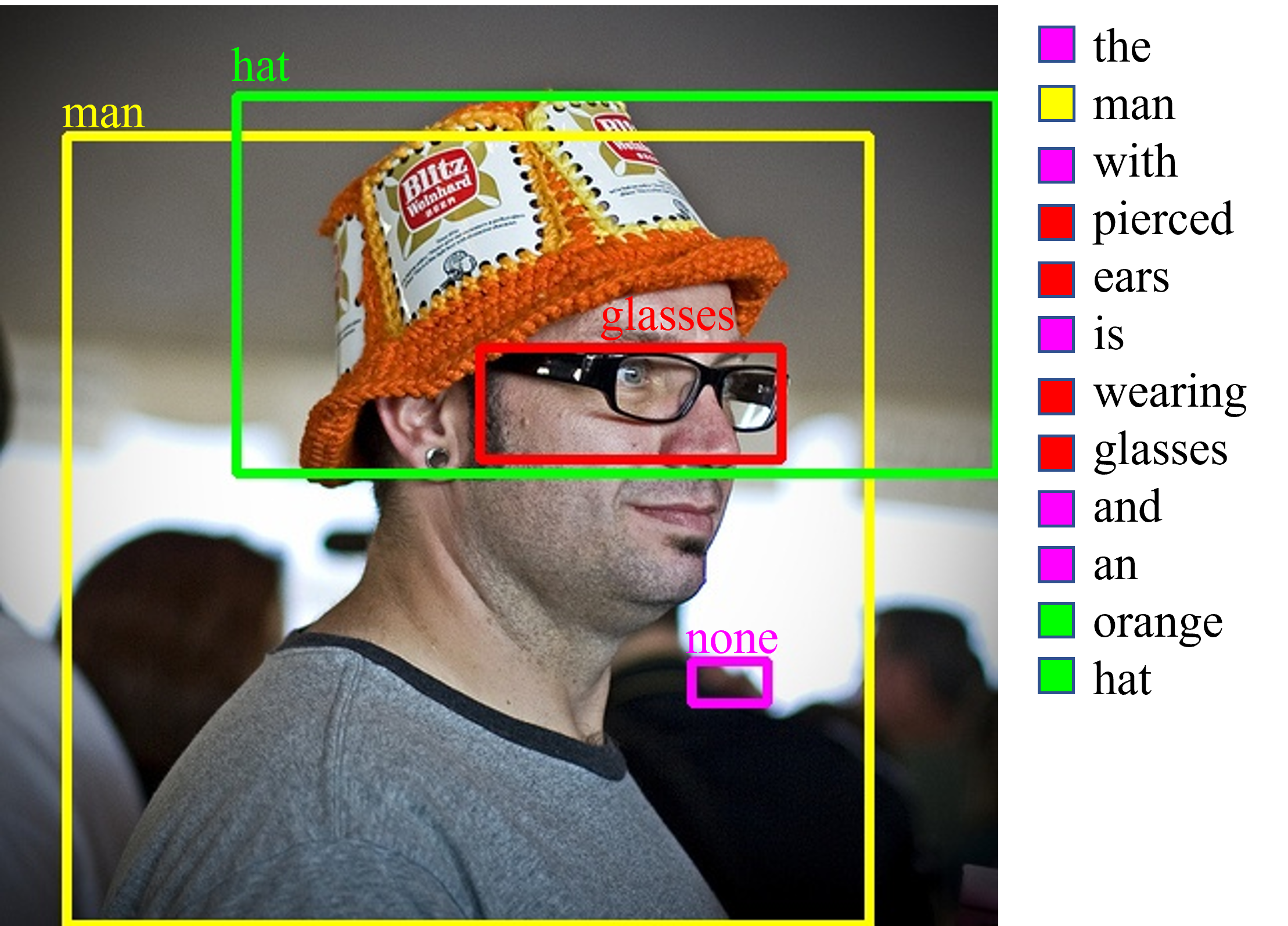}}
    \subfigure[Proposed regions.]{\includegraphics[width=0.255\textwidth]{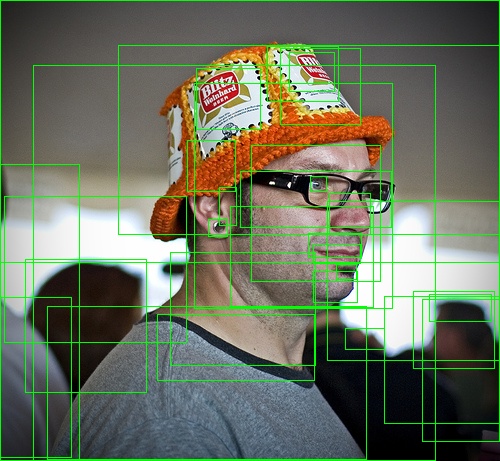}}
    \caption{Result visualization. In the figure(a)(b), the word on each proposed region represent the prominent word selected during cross-modal semantic fusion for that region. 
    The color to the left of each word on the right side of the figure indicates the color of the prominent region selected when performing cross-modal fine-grained alignment. 
    Figure (c) displays all the proposed regions extracted by Faster RCNN.
    }
    \label{fig3}
\end{figure*}

\subsection{Result Visualization}
The following visualizations present our experimental results to illustrate the effectiveness of our proposed method more clearly. 
Figures \ref{fig3} depict the fine-grained alignment results between the displayed images and the text description ``The man with pierced ears is wearing glasses and an orange hat."
In figure \ref{fig3}(a), the alignment results show how each word aligns with various regions without cross-modal semantic fusion($V$ in section \ref{33}), 
while figure \ref{fig3}(b) shows the alignment results when each word aligns with regions that have undergone cross-modal semantic fusion($V^*$ in section \ref{33}).

Our experimental results show that during cross-modal information fusion, only 4 prominent words are fused with all the proposed regions. 
These words are ``man", ``glasses," ``hat," and ``none," where ``none" signifies that the region is not fused with any word. 
Among all the proposed regions(shows in figure \ref{fig3}(c)), we only display the regions that are aligned with words. 
In cross-modal fine-grained alignment, out of over 30 proposed regions, only the 4 regions shown in the figure are aligned with words, 
while the rest of the regions are not involved in the alignment. This demonstrates the success of our method in highlighting prominent regions and words.

Before cross-modal semantic fusion, semantic alignment in the text can be divided into four groups: ``the man is," ``with pierced ears wearing glasses," ``and," and ``an orange hat." 
The semantic grouping seems reasonable, but ``ears" and ``glasses" are not separated, possibly due to the absence of regions related to ``ears" in the proposed boxes. 
After performing cross-modal semantic fusion, the yellow region which are relatively less significant compared to the other three regions is removed. 
The purple region participates in the alignment, and its similarity with all words becomes negative. 
This leads to an increase in the similarity score for prominent regions and a relative decrease in the similarity score for irrelevant regions.
For the words, the words ``the," ``with," ``is," ``and," and ``an" align with irrelevant regions in the image, 
while the regions corresponding to other words remain unchanged. 
This diminishes the impact of interchangeable words. 
After combining the similarity between two types of region representations, 
the similarity of prominent words relatively increases, while the similarity of other words decreases.

\section{Conclusion}
In this paper, we propose a cross-modal prominent fragment enhancement alignment method for fine-grained alignment in image-text retrieval. 
Our model aims to (1) discover prominent fragments in images and text, and enhance the influence of these prominent fragments in fine-grained alignment. 
(2) Explore the semantic and spatial relationships among different regions in images and the semantic relationships among different words in text. 
Conducted ablation experiments and visual results demonstrate the effectiveness of our proposed method. 
Extensive comparative experiments on the Flickr30K and MS-COCO datasets show that our model achieves state-of-the-art performance.


\begin{thebibliography}{00}
\bibitem{b1} Andrea Frome, Greg S Corrado, Jon Shlens, Samy Bengio, Jeff Dean, Marc’Aurelio Ranzato, and Tomas Mikolov. De- vise: A deep visual-semantic embedding model. In NeurIPS, pages 2121–2129, 2013.
\bibitem{b2} Fartash Faghri, David J Fleet, Jamie Ryan Kiros, and Sanja Fidler. Vse++: Improving visual-semantic embeddings with hard negatives. In BMVC, 2018.
\bibitem{b3} Jiacheng Chen, Hexiang Hu, Hao Wu, Yuning Jiang, and Changhu Wang. Learning the best pooling strategy for vi- sual semantic embedding. In CVPR, pages 15789–15798, 2021.
\bibitem{b4} Kunpeng Li, Yulun Zhang, Kai Li, Yuanyuan Li, and Yun Fu. Visual semantic reasoning for image-text matching. In ICCV, pages 4654–4662, 2019.
\bibitem{b5} Peter Anderson, Xiaodong He, Chris Buehler, Damien Teney, Mark Johnson, Stephen Gould, and Lei Zhang. Bottom-up and top-down attention for image captioning and visual question answering. In CVPR, pages 6077–6086, 2018.
\bibitem{b6} Ashish Vaswani, Noam Shazeer, Niki Parmar, Jakob Uszkoreit, Llion Jones, Aidan N Gomez, Łukasz Kaiser, and Illia Polosukhin. Attention is all you need. In NeurIPS, pages 5998–6008, 2017.
\bibitem{b7} Kuang-Huei Lee, Xi Chen, Gang Hua, Houdong Hu, and Xiaodong He. Stacked cross attention for image-text matching. In ECCV, pages 201–216, 2018.
\bibitem{b8} Haiwen Diao, Ying Zhang, Lin Ma, and Huchuan Lu. Similarity reasoning and filtration for image-text matching. In AAAI, volume 35, pages 1218–1226, 2021.
\bibitem{b9} Kun Zhang, Zhendong Mao, Quan Wang, and Yongdong Zhang. Negative-aware attention framework for image-text matching. In CVPR, pages 15661–15670, 2022.
\bibitem{b10} Y. Liu, H. Liu, H. Wang, F. Meng and M. Liu, "BCAN: Bidirectional Correct Attention Network for Cross-Modal Retrieval," in IEEE Transactions on Neural Networks and Learning Systems, doi: 10.1109/TNNLS.2023.3276796.
\bibitem{b11} H. Chen, G. Ding, X. Liu, Z. Lin, J. Liu and J. Han, "IMRAM: Iterative Matching With Recurrent Attention Memory for Cross-Modal Image-Text Retrieval," 2020 IEEE/CVF Conference on Computer Vision and Pattern Recognition (CVPR), Seattle, WA, USA, 2020, pp. 12652-12660, doi: 10.1109/CVPR42600.2020.01267.
\bibitem{b12} Yuhao Cheng, Xiaoguang Zhu, Jiuchao Qian, Fei Wen, and Peilin Liu. 2022. Cross-modal Graph Matching Network for Image-text Retrieval. ACM Trans. Multimedia Comput. Commun. Appl. 18, 4, Article 95 (November 2022), 23 pages. https://doi.org/10.1145/3499027
\bibitem{b13} C. Liu, Z. Mao, T. Zhang, H. Xie, B. Wang and Y. Zhang, "Graph Structured Network for Image-Text Matching," 2020 IEEE/CVF Conference on Computer Vision and Pattern Recognition (CVPR), Seattle, WA, USA, 2020, pp. 10918-10927, doi: 10.1109/CVPR42600.2020.01093.
\bibitem{b14} J. Johnson et al., "Image retrieval using scene graphs," 2015 IEEE Conference on Computer Vision and Pattern Recognition (CVPR), Boston, MA, USA, 2015, pp. 3668-3678, doi: 10.1109/CVPR.2015.7298990.
\bibitem{b15} Nicola Messina, Giuseppe Amato, Andrea Esuli, Fabrizio Falchi, Claudio Gennaro, and Stéphane Marchand-Maillet. 2021. Fine-Grained Visual Textual Alignment for Cross-Modal Retrieval Using Transformer Encoders. ACM Trans. Multimedia Comput. Commun. Appl. 17, 4, Article 128 (November 2021), 23 pages. https://doi.org/10.1145/3451390
\bibitem{b16} Z. Fu, Z. Mao, Y. Song and Y. Zhang, "Learning Semantic Relationship among Instances for Image-Text Matching," 2023 IEEE/CVF Conference on Computer Vision and Pattern Recognition (CVPR), Vancouver, BC, Canada, 2023, pp. 15159-15168, doi: 10.1109/CVPR52729.2023.01455.
\bibitem{b17} X. Ge, F. Chen, S. Xu, F. Tao and J. M. Jose, "Cross-modal Semantic Enhanced Interaction for Image-Sentence Retrieval," 2023 IEEE/CVF Winter Conference on Applications of Computer Vision (WACV), Waikoloa, HI, USA, 2023, pp. 1022-1031, doi: 10.1109/WACV56688.2023.00108.
\bibitem{b18} Y. Chen et al., "More Than Just Attention: Improving Cross-Modal Attentions with Contrastive Constraints for Image-Text Matching," 2023 IEEE/CVF Winter Conference on Applications of Computer Vision (WACV), Waikoloa, HI, USA, 2023, pp. 4421-4429, doi: 10.1109/WACV56688.2023.00441.
\bibitem{b19} Z. Pan, F. Wu and B. Zhang, "Fine-grained Image-text Matching by Cross-modal Hard Aligning Network," 2023 IEEE/CVF Conference on Computer Vision and Pattern Recognition (CVPR), Vancouver, BC, Canada, 2023, pp. 19275-19284, doi: 10.1109/CVPR52729.2023.01847.
\bibitem{b20} Liwei Wang, Yin Li, and Svetlana Lazebnik. Learning deep structure-preserving image-text embeddings. In CVPR, pages 5005–5013, 2016.
\bibitem{b21} Ryan Kiros, Ruslan Salakhutdinov, and Richard S Zemel. Unifying visual-semantic embeddings with multimodal neural language models. Trans. Assoc. Comput. Linguist, 2015.
\bibitem{b22} Jacob Devlin, Ming-Wei Chang, Kenton Lee, and Kristina Toutanova. Bert: Pre-training of deep bidirectional transformers for language understanding. ACL, 2018.
\bibitem{b23} Xuri Ge, Fuhai Chen, Joemon M. Jose, Zhilong Ji, Zhongqin Wu, and Xiao Liu. 2021. Structured Multi-modal Feature Embedding and Alignment for Image-Sentence Retrieval. In Proceedings of the 29th ACM International Conference on Multimedia (MM '21). Association for Computing Machinery, New York, NY, USA, 5185–5193. https://doi.org/10.1145/3474085.3475634
\bibitem{b24} Chunxiao Liu, Zhendong Mao, An-An Liu, Tianzhu Zhang, Bin Wang, and Yongdong Zhang. 2019. Focus Your Attention: A Bidirectional Focal Attention Network for Image-Text Matching. In Proceedings of the 27th ACM International Conference on Multimedia (MM '19). Association for Computing Machinery, New York, NY, USA, 3–11. https://doi.org/10.1145/3343031.3350869
\bibitem{b25} F. Zhan et al., "Multimodal Image Synthesis and Editing: A Survey and Taxonomy," in IEEE Transactions on Pattern Analysis and Machine Intelligence, doi: 10.1109/TPAMI.2023.3305243.
\bibitem{b26} S. Antol et al., "VQA: Visual Question Answering," 2015 IEEE International Conference on Computer Vision (ICCV), Santiago, Chile, 2015, pp. 2425-2433, doi: 10.1109/ICCV.2015.279.
\bibitem{b27} L. Qiao and W. Hu, "A Survey of Deep learning-based Image caption," 2022 2nd International Conference on Computer Science, Electronic Information Engineering and Intelligent Control Technology (CEI), Nanjing, China, 2022, pp. 120-123, doi: 10.1109/CEI57409.2022.9950180.
\bibitem{b28} S. Long, S. C. Han, X. Wan and J. Poon, "GraDual: Graph-based Dual-modal Representation for Image-Text Matching," 2022 IEEE/CVF Winter Conference on Applications of Computer Vision (WACV), Waikoloa, HI, USA, 2022, pp. 2463-2472, doi: 10.1109/WACV51458.2022.00252.
\bibitem{b29} Christopher D. Manning, Mihai Surdeanu, John Bauer, Jenny Rose Finkel, Steven Bethard, and David McClosky. 2014. The Stanford CoreNLP natural language processing toolkit. In Proceedings of 52nd Annual Meeting of the Association for Computational Linguistics: System Demonstrations. 55–60.
\bibitem{b30} K. Li, Y. Zhang, K. Li, Y. Li and Y. Fu, "Image-Text Embedding Learning via Visual and Textual Semantic Reasoning," in IEEE Transactions on Pattern Analysis and Machine Intelligence, vol. 45, no. 1, pp. 641-656, 1 Jan. 2023, doi: 10.1109/TPAMI.2022.3148470.
\bibitem{b31} T.-Y. Lin et al., “Microsoft COCO: Common objects in context,” in Proc. ECCV. Cham, Switzerland: Springer, 2014, pp. 740–755.
\bibitem{b32} B. A. Plummer, L. Wang, C. M. Cervantes, J. C. Caicedo, J. Hockenmaier, and S. Lazebnik, “Flickr30k entities: Collecting region-to-phrase correspondences for richer image-to-sentence models,” in Proc. IEEE Int. Conf. Comput. Vis. (ICCV), Dec. 2015, pp. 2641–2649.
\bibitem{b33} Qi Zhang, Zhen Lei, Zhaoxiang Zhang, and Stan Z Li. Context-aware attention network for image-text retrieval. In CVPR, pages 3536–3545, 2020.
\end{thebibliography}
\end{document}